%% file: main.tex
\documentclass[10pt,twocolumn,letterpaper]{article}

\usepackage[pagenumbers]{iccv} 
\usepackage[accsupp]{axessibility}  

\usepackage{multirow}
\usepackage{booktabs}
\usepackage{graphicx}
\usepackage{pifont}

\definecolor{iccvblue}{rgb}{0.21,0.49,0.74}
\definecolor{lightpurple}{rgb}{0.902, 0.863, 0.922}
\definecolor{lightgreen}{rgb}{0.863, 0.902, 0.863}
\definecolor{lightpink}{rgb}{0.980, 0.941, 0.941}
\usepackage[pagebackref,breaklinks,colorlinks,allcolors=iccvblue]{hyperref}

\def\paperID{***} 
\def\confName{ICCV}
\def\confYear{2025}

\title{Fine-structure Preserved Real-world Image Super-resolution via \\ Transfer VAE Training}

\author{Qiaosi Yi$^{1,2}$,  Shuai Li$^{1}$,  Rongyuan Wu$^{1,2}$, Lingchen Sun$^{1,2}$, Yuhui Wu$^{1,2}$, Lei Zhang$^{1,2}$\thanks{Corresponding author. This research is supported by the PolyU-OPPO Joint Innovative Research Center.} \\
{$^{1}$The Hong Kong Polytechnic University \qquad $^{2}$OPPO Research Institute} \\
{\tt\small  qiaosiyijoyies@gmail.com, cslzhang@comp.polyu.edu.hk}  \\ 
{\tt\small \{novak.li, rong-yuan.wu, ling-chen.sun, yuhui.wu\}@connect.polyu.hk} 
}

\begin{document}
\maketitle
\input{sec/0_abstract}    
\input{sec/1_intro}

\input{sec/2_relatedwork}
\input{sec/3_method}
\input{sec/4_exp}
\input{sec/5_con}

{
    \small
    \bibliographystyle{ieeenat_fullname}
    \bibliography{main}
}


\end{document}

%% file: sec/0_abstract.tex
\begin{abstract}
Impressive results on real-world image super-resolution (Real-ISR) have been achieved by employing pre-trained stable diffusion~(SD) models. However, one critical issue of such methods lies in their poor reconstruction of image fine structures, such as small characters and textures, due to the aggressive resolution reduction of the VAE (\eg, 8$\times$ downsampling) in the SD model. One solution is to employ a VAE with a lower downsampling rate for diffusion; however, adapting its latent features with the pre-trained UNet while mitigating the increased computational cost poses new challenges. 
To address these issues, we propose a Transfer VAE Training (TVT) strategy to transfer the 8$\times$ downsampled VAE into a 4$\times$ one while adapting to the pre-trained UNet. Specifically, we first train a 4$\times$ decoder based on the output features of the original VAE encoder, then train a 4$\times$ encoder while keeping the newly trained decoder fixed. Such a TVT strategy aligns the new encoder-decoder pair with the original VAE latent space while enhancing image fine details.
Additionally, we introduce a compact VAE and compute-efficient UNet by optimizing their network architectures, reducing the computational cost while capturing high-resolution fine-scale features.
Experimental results demonstrate that our TVT method significantly improves fine-structure preservation, which is often compromised by other SD-based methods, while requiring fewer FLOPs than state-of-the-art one-step diffusion models. The official code can be found at https://github.com/Joyies/TVT.
\end{abstract}

%% file: sec/1_intro.tex
\section{Introduction}
\label{sec:intro}

Aiming at reproducing the high-resolution (HR) image from a low-resolution (LR) input, image super-resolution (ISR) has been extensively studied in past decades \cite{edsr,rcan,rdn,san, srgan, wang2018esrgan, liang2022details, peng2024efficient, kong2022reflash, zhang2021designing, wang2021real,wei2025perceive,kong2021classsr}. 
Despite the remarkable progresses in network architecture design \cite{edsr,rcan,rdn,san,liang2021swinir,chen2025generalized} and training losses \cite{johnson2016perceptual,ssim,goodfellow2014generative}, real-world ISR (Real-ISR) remains a challenging problem due to the complex, unknown, and entangled image degradations (\eg, sensor noise, motion blur, compression artifacts). 
Many researches~\cite{srgan, wang2018esrgan, liang2022details, zhang2021designing, wang2021real, sun2024perception} have focused on the use of generative adversarial networks (GANs)~\cite{goodfellow2014generative} to generate photorealistic textures and details, but the generative prior of GAN is limited and the adversarial training process will introduce many undesirable visual artifacts in the super-resolved (SR) images~\cite{liang2022details}.

\begin{figure}
	\centering
	\includegraphics[width=0.9\linewidth]
	{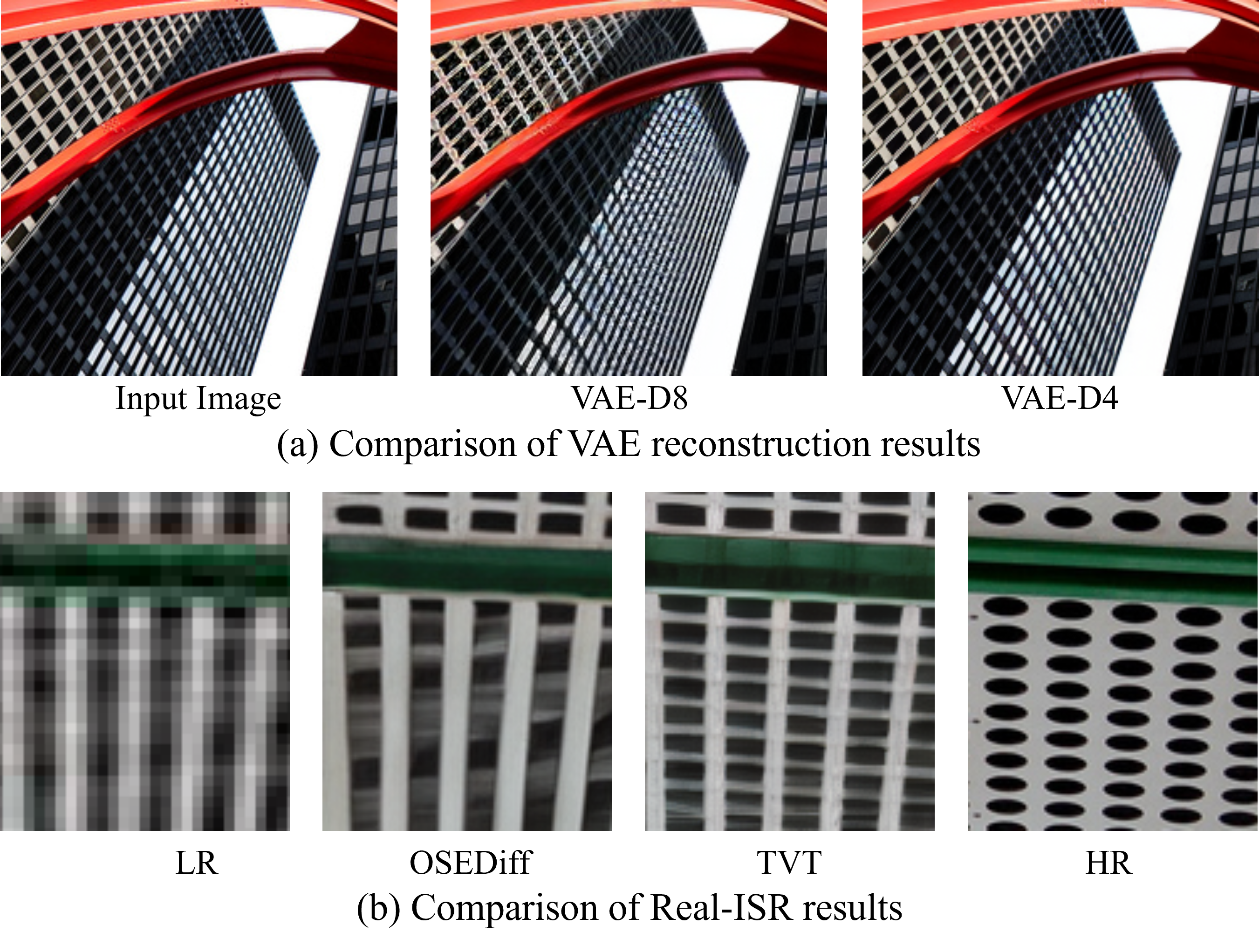}
        \vspace{-4mm}
	\caption{(a) Comparison of the reconstruction results of original VAE with 8$\times$  resolution downsampling (VAE-D8) and our adapted VAE with 4$\times$ resolution downsampling (VAE-D4). (b) Comparison of the Real-ISR results by the state-of-the-art one step diffusion based method OSEDiff \cite{wu2024one} and our method with Transfer VAE Training (TVT). Please zoom in for a better view.}
	\label{fig:introimage1}
    \vspace{-4mm}
\end{figure}

Recent advances in large-scale text-to-image (T2I) diffusion models~\cite{saharia2022photorealistic,rombach2022high} have revolutionized the image generation task, as well as many down-streaming tasks such as Real-ISR. Based on pre-trained stable diffusion (SD) models~\cite{rombach2022high}, many Real-ISR methods~\cite{wang2023exploiting,yang2023pixel,lin2023diffbir,wu2023seesr,sun2023improving,yu2024scaling} have been proposed to improve the realistic details of ISR outputs. Earlier works such as StableSR~\cite{wang2023exploiting}, PASD \cite{yang2023pixel} and SeeSR~\cite{wu2023seesr} take the LR input as the control signal and utilize ControlNet~\cite{zhang2023adding} to guide the ISR process. However, these methods need to use random noise as the seed and perform multiple denoising steps to achieve the results. Very recently, one-step SD-based Real-ISR methods~\cite{wu2024one, zhang2024degradation, sun2024pixel}
have been developed, which directly takes the LR image as the input and uses score distillation techniques to finetune the UNet, significantly reducing the computational cost.
Albeit the visually appealing results achieved by existing SD-based Real-ISR methods, one well-known and critical issue of them lies in their difficulties in reproducing image fine-structures, such as small-scale details and textures, as illustrated in Fig.~\ref{fig:introimage1}(b). This is essentially caused by the information loss in the variational auto-encoder (VAE) employed in the SD model. Specifically, the VAE used in SD 2.1-base \cite{sd} compresses a $512\times512\times3$ color image into a $64\times64\times4$ shaped latent feature through $8\times$ spatial compression, causing irreversible loss of fine-scale details (see Fig.~\ref{fig:introimage1}(a)) that cannot be recovered by the diffusion process. 

To alleviate the information loss caused by the reduction in VAE resolution, a natural remedy is to employ a VAE with a lower compression ratio (\eg, $4\times$ downsampling) to retain more fine-grained details in the latent space. However, this will introduce new challenges. First, how to adapt the latent features of the new VAE to the pre-trained UNet in SD is non-trivial, as a mismatched latent space can hinder the generative capability of the pre-trained diffusion. Second, the increased resolution of the latent features will significantly increase the computational cost of the diffusion process, making it highly expensive for the Real-ISR task.

To address these issues, we propose a Transfer VAE Training (TVT) approach for fine-structure preserved Real-ISR. 
Specifically, we propose a TVT strategy to systematically adapt the original $8\times$ downsampled VAE~(VAE-D8) into a $4\times$ downsampled VAE (VAE-D4) while maintaining its compatibility with the pre-trained denoising UNet. Note that the latent space of the adapted VAE should closely resemble the original one to exploit the pre-trained UNet’s diffusion prior. To this end, we first train a compact VAE-D4 decoder conditioned on the VAE-D8 encoder's outputs so that the decoder can learn to reconstruct HR images with enhanced details while aligning with the original latent distribution. Then we train a compact VAE-D4 encoder by fixing the newly trained decoder, further stabilizing the latent space transition. This two-stage TVT strategy enables the seamless integration of our adapted VAE-D4 into the SD framework without retraining the UNet from scratch. 
In addition, to reduce the computational overhead caused by high-resolution latent features, we design a compute-efficient UNet to effectively process fine-scale features. Unlike previous work \cite{chen2024adversarial} that decreases channel dimensions, we replicate the first and last layers of the pre-trained UNet and add them to the UNet. After fine-tuning, the replicated layers could handle $128\times128$ resolution features, while the original UNet continues to process $64\times64$ latent features. Such a design can reduce the FLOPs significantly while preserving the diffusion priors of the pre-trained UNet.

Extensive experiments demonstrate that our TVT method outperforms state-of-the-art SD-based Real-ISR methods in preserving image fine structures (see Fig.~\ref{fig:introimage1}(b)). Additionally, our method achieves lower FLOPs than recent one-step diffusion-based methods, thanks to the compact VAE and efficient UNet branch.

%% file: sec/2_relatedwork.tex
\section{Related Work}
\label{sec:relatedwork}

\textbf{Real-world image super-resolution}.
Early ISR methods~\cite{edsr,rcan,li2018multi,rdn,san,chen2021pre,dai2021feedback,liang2021swinir,zhang2022efficient,chen2023activating, dat,chen2025generalized} primarily employ pixel-level loss functions~(\eg, $L_1$ and $L_2$ losses) to enhance the fidelity of ISR results utilizing synthetic LR-HR pairs generated via bicubic downsampling. 
However, these methods often fail to handle images with complex degradations in real-world scenes~\cite{wang2021real} and result in over-smoothed ISR outputs~\cite{dong2014learning, gu2019blind}. To alleviate the limitation of synthetic datasets, some researchers have used long-short camera focal lenses to collect real-world paired datasets~\cite{realsr, drealsr, liang2024ntire}, such as RealSR and DrealSR datasets. Meanwhile, some researchers proposed to use more complex degradation pipelines to simulate degradation in real scenarios~\cite{zhang2021designing,wang2021real}. For example, BSRGAN~\cite{zhang2021designing} synthesizes realistic LR-HR pairs by randomly shuffling basic degradation operators (\eg, blur, noise, downsampling), while RealESRGAN~\cite{wang2021real} introduces high-order degradations to simulate real-world distortions. In addition, some perceptual loss functions~\cite{johnson2016perceptual,wang2004image} and GAN-based methods~\cite{wang2018esrgan, zhang2021designing, chen2022femasr, liang2022efficient} have been introduced to enhance the photorealistic appearance of the ISR results. Despite the success in generating realistic details and textures, GAN-based methods often produce unnatural artifacts~\cite{liang2022details,xie2023desra,wang2021real, sun2024perception} due to training instability. Furthermore, the generative prior of GAN-based methods is limited, hindering their effectiveness in Real-ISR.

\begin{figure*}
	\centering
	\includegraphics[width=0.97\linewidth]
	{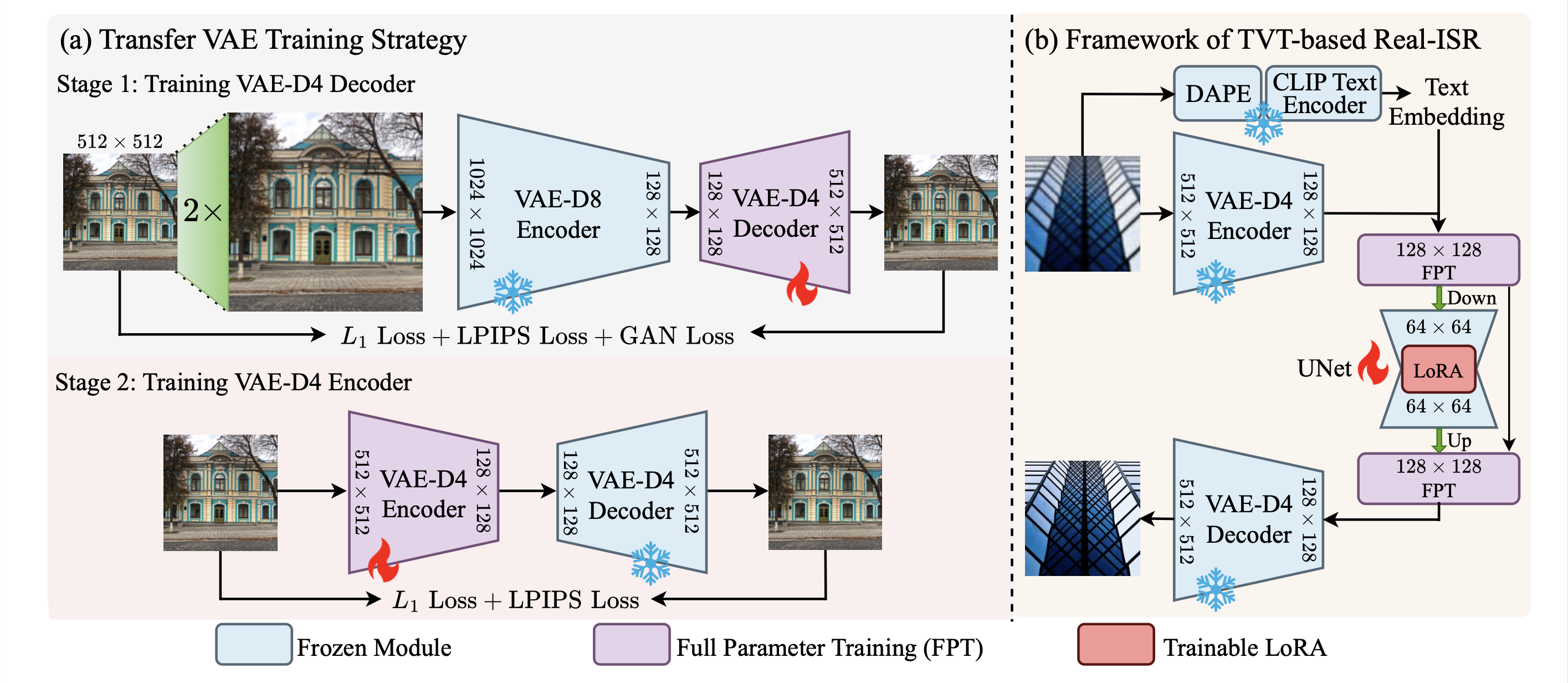}
        \vspace{-4mm}
	\caption{(a) Illustration of our transfer VAE training strategy. We first train a VAE-D4 decoder conditioned on the original VAE-D8's encoder output by using $L_1$ loss, LPIPS loss, and GAN loss, ensuring that the decoder can accurately reconstruct the VAE-D8's latent features into high-quality images. Then we train the VAE-D4 encoder by fixing the newly trained decoder, ensuring that the encoder can match the desired latent space. Finally, the trained VAE-D4 can be seamlessly adapted to the pre-trained SD-UNet while preserving fine structures. (b) With the adapted VAE-D4, we replicate the denoising SD-UNet's first and last layers which are integrated into the UNet to process $128\times128$ features. Meanwhile, the original SD-UNet is employed to process $64\times64$ features. Note that only the replicated layers are fully fine-tuned, while the denoising UNet is updated via LoRA finetuning.}
	\label{fig:model}
    \vspace{-4mm}
\end{figure*}

\textbf{Diffusion model based super-resolution}. 
Diffusion models have emerged as a powerful framework for image generation~\cite{ho2020denoising, song2020score, dhariwal2021diffusion, zhang2023adding, mou2024t2i,peng2024towards,xie2025dnaedit,wu2025insvie}, and they have been successfully used in the Real-ISR task. Early approaches~\cite{kawar2022denoising} adapt pre-trained diffusion models by adjusting their reverse transition processes to restore clear images; however, these methods fail to handle real-world degradations like noise and blur. Some works train diffusion models from scratch~\cite{kawar2021snips, kawar2022denoising, wang2022zero, yue2023ResShift,mei2024codi} on paired LR-HR datasets, but their performance is constrained by the diversity and scale of available datasets. With the great success of SD~\cite{sd} in the T2I task, researchers propose many SD-based Real-ISR methods \cite{wang2023exploiting, ai2024dreamclear, qu2024xpsr, lin2023diffbir, yang2023pixel, wu2023seesr, yu2024scaling, wu2024one, sun2024pixel, mei2025power}, which integrate ControlNet~\cite{zhang2023adding}, T2I-adapter~\cite{mou2024t2i}, restoration modules, and semantic information extractors in SD to achieve visually pleasing results.  
However, these methods suffer from instability taking random noise as input, and require dozens of diffusion steps to process an image. To accelerate the diffusion process, 
OSEDiff~\cite{wu2024one} directly takes the LR image as input without using random noise, employs the lightweight LoRA adapters~\cite{lora} and the VSD loss~\cite{wang2024prolificdreamer, yin2024one} to compress the multiple diffusion steps into one-step process. S3Diff~\cite{zhang2024degradation} implements a one-step Real-ISR method using a degradation-guided LoRA adapter, with GAN loss for optimization. Despite the visually appealing results, these methods fail to recover fine image structures such as small text, faces, and textures due to the information loss caused by the $8\times$ spatial compression of the VAE encoder.

%% file: sec/3_method.tex
\section{Methodology}

\subsection{Motivation}

To enhance the training efficiency for text-to-image (T2I) tasks, SD \cite{rombach2022high} employs a VAE-D8 to compress high-dimensional images into low-dimensional latent features with $8\times$ spatial downsampling (\eg, $512\times512\times3$ → $64\times64\times4$). While this compression reduces the training cost and alleviates optimization difficulties for the T2I task, it irreversibly loses high-frequency details, which poses  significant challenges for the Real-ISR task, where fine-grained structure is crucial. Recent attempts~\cite{esser2024scaling,flux2024} alleviate the aforementioned limitations by increasing latent channels. However, their diffusion models are trained from scratch, which is computationally expensive.

To mitigate the fine-structure preservation limitation of VAE-D8 without re-training the SD model from scratch, we propose a Transfer VAE Training (TVT) strategy, as illustrated in Fig.~\ref{fig:model}(a). The key idea is to introduces a $4\times$ downsampled VAE (VAE-D4) to preserve structural integrity while maintaining its compatibility with pre-trained SD UNet. Our TVT consists of two stages: first training a compact VAE-D4 decoder using the outputs of the VAE-D8 encoder, then training a compact VAE-D4 encoder by keeping the VAE-D4 decoder fixed. Through this two-stage training, the adapted VAE-D4 can not only achieve good latent space alignment with the original VAE-D8 but also preserve more fine-scale structures of the original image. 

With the adapted VAE-D4, we consequently propose a framework for TVT-based Real-ISR, as shown in Fig.~\ref{fig:model}(b). To efficiently process the $128 \times 128$ latent features of VAE-D4, we replicate the first and last layers of the pre-trained UNet and then integrate them back into the original UNet. These replicated layers are fully fine-tuned to process the $128 \times 128$ latent features while maintaining the skip connections between them. Meanwhile, the original UNet is fine-tuned using LoRA to handle $64 \times 64$ latent features. Our design reduces much the FLOPs while preserving the diffusion priors of the pre-trained UNet.

\subsection{Transfer VAE Training}

\begin{figure}
	\centering
	\includegraphics[width=1\linewidth]
	{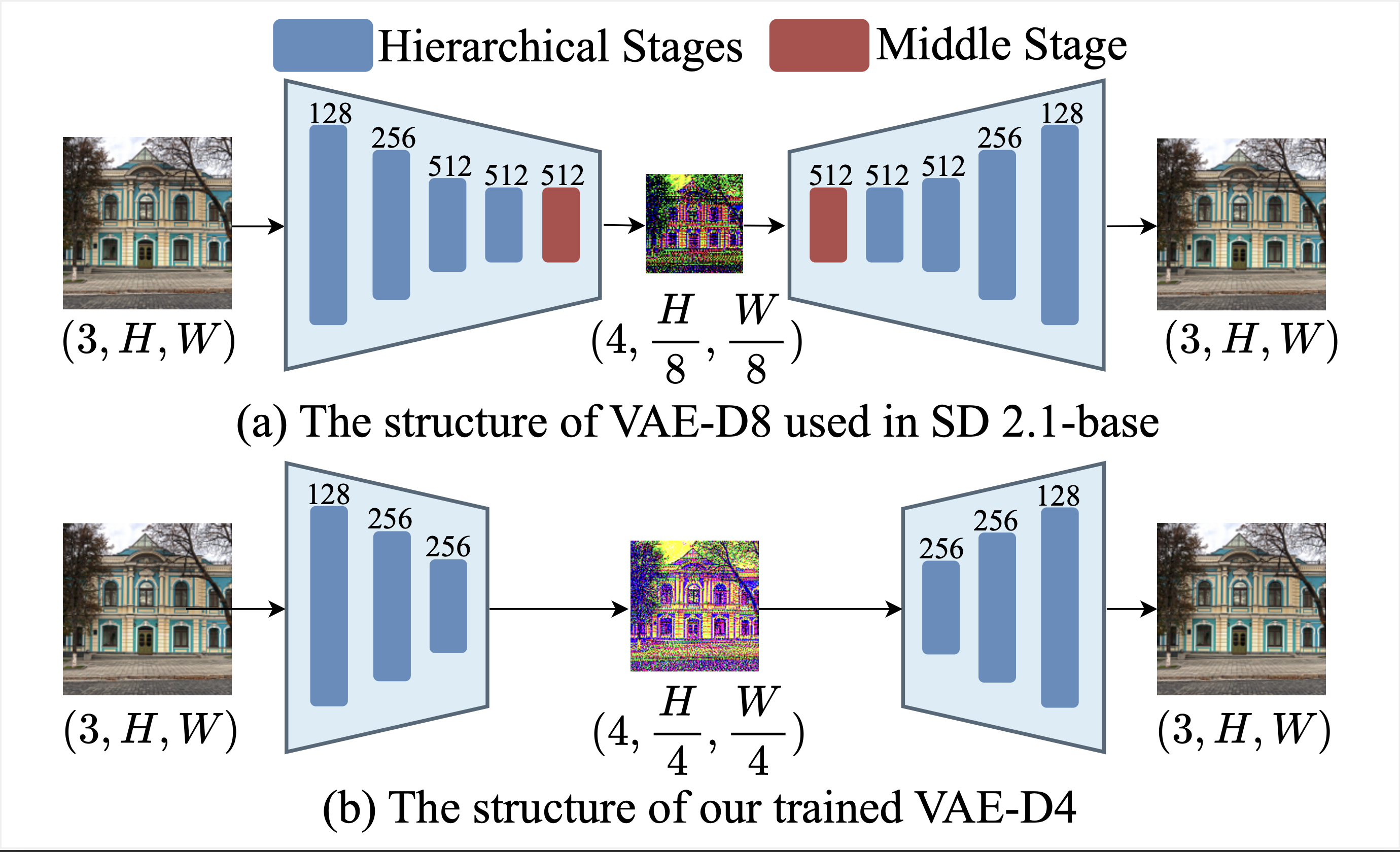}
        \vspace{-6mm}
	\caption{The structure of (a) VAE-D8 used in SD 2.1-base and (b) our proposed compact VAE-D4.}
	\label{fig:vae}
    \vspace{-4mm}
\end{figure}

\textbf{Compact VAE-D4 Design}. The original VAE-D8 architecture consists of 5 stages, denoted by $\{S_i\}_{i=1}^5$, as shown in Fig.~\ref{fig:vae}(a). $\{S_i\}_{i=1}^4$ utilizes ResNet blocks~\cite{he2016deep} (ResBlocks) for feature extraction with increasing channel dimensions of 128, 256, 512, and 512. $S_5$ is a middle stage, which is introduced to integrate self-attention modules into ResBlocks for enhanced feature representation learning. While a naive design of our VAE-D4 is to adopt the structure of VAE-D8 (\eg, 4 hierarchical stages plus a middle stage), such a configuration incurs excessive computational costs. Considering that reducing the downsampling ratio can better preserve spatial structural details, it is unnecessary to introduce a complex VAE model. Therefore, we propose a compact VAE-D4 design to reduce redundancy of features, as shown in Fig.~\ref{fig:vae}(b). Specifically, we reduce the number of stages from 5 to 3 by removing a hierarchical stage and the middle stage, then prune the number of channels to 128/256/256. Finally, we achieve a reduction of 81.89\% parameters and 38.89\% FLOPs while maintaining the reconstruction fidelity. 

Based on the compact VAE-D4, we propose a two-stage TVT strategy to align its latent space with the original VAE-D8. First, we train the VAE-D4 decoder to ensure that it can reconstruct clear images from the latent space of VAE-D8; then, we train the VAE-D4 encoder to generate latent features obeying the distribution of the adapted decoder.  

\noindent
\textbf{Training VAE-D4 Decoder.} As shown in Fig.~\ref{fig:model}(a), we train the VAE-D4 decoder conditioned on the latent features of the VAE-D8 encoder. However, the different compression ratios of VAE-D8 and VAE-D4 introduce a resolution mismatch: when an input image $I\in\mathbb{R}^{h \times w \times 3}$ is encoded by VAE-D8, and the resulting latent feature $z \in\mathbb{R}^{\frac{h}{8} \times \frac{w}{8} \times 4}$ is decoded by VAE-D4 to obtain the output image $O\in\mathbb{R}^{\frac{h}{2} \times \frac{w}{2} \times 3}$, whose spatial resolution will be half of the original image $I$. To resolve this issue, we first upsample $I$ by $2\times$ to $\tilde{I} \in \mathbb{R}^{2h \times 2w \times 3}$, and then encode $\tilde{I}$ with VAE-D8 to yield a latent feature $\tilde{z} \in \mathbb{R}^{\frac{H}{4} \times \frac{W}{4} \times C} $, which can be decoded to $O \in \mathbb{R}^{H \times W \times 3}$ by our VAE-D4 decoder:
\begin{equation}
    O = D_4(E_8(I\uparrow^2)),
\end{equation}
where $D_4$ and $E_8$ denote the VAE-D4 decoder and VAE-D8 encoder, respectively, and $\uparrow^2$ denotes the upsampling operator with scale factor 2. Similar to SD~\cite{sd}, we use $L_1$ loss, LPIPS loss and GAN loss to train the VAE-D4 decoder to ensure that it can generate realistic details: 
\begin{equation}
    \mathcal{L}_{D} = L_{rec}(I,O) + \lambda_{D} L_{GAN}(I,O).
\end{equation}
where $L_{rec}=L_1+L_{LPIPS}$, $\lambda_{D} =\frac{\nabla[L_{rec}]}{\nabla[L_{GAN}]+10^{-6}}$ and $\nabla[\cdot]$ is the gradient of the last layer of decoder.

\noindent
\textbf{Training VAE-D4 Encoder.} After training the VAE-D4 decoder, we fix it to train the corresponding VAE-D4 encoder. As shown in Fig.~\ref{fig:model}(a) stage 2, the input image $I\in\mathbb{R}^{h \times w \times 3}$ will be encoded by VAE-D4 encoder to obtain the latent feature $z \in\mathbb{R}^{\frac{h}{4} \times \frac{w}{4} \times 4}$, which is then decoded by the VAE-D4 decoder back into an image $O\in\mathbb{R}^{h \times w \times 3}$: 
\begin{equation}
O = D_4(E_4(I)),
\end{equation}
where $E_4$ and $D_4$ are the encoder and decoder of VAE-D4. We calculate the loss between $I$ and $O$ to optimize the parameters of encoder $E_4$. We empirically eliminate the GAN loss during the training of the encoder as we observe that incorporating the GAN loss leads to training instability. The $L_1$ and LPIPS losses are used to train the $E_4$:
\begin{equation}
    \mathcal{L}_{E} = L_1(I,O) + L_{LPIPS}(I,O).
\end{equation}

\begin{table}
		\caption{Reconstruction performance comparison between VAE-D8 and VAE-D4 on Urban100 \cite{huang2015single} dataset.}
         \vspace{-3mm}
    \centering
	\resizebox{\linewidth}{!}{
		\begin{tabular}{c|ccccc}
			\toprule
			VAE & FLOPs & Para. & PSNR↑ & SSIM↑ & LPIPS↓ \\ \midrule
			VAE-D8~\cite{sd} & 1.8 T & 83.7 M  &25.29 & 0.7656 & 0.0659  \\
       Our VAE-D4 & \textbf{1.1 T} & \textbf{15.2 M} & \textbf{32.35} & \textbf{0.9290} & \textbf{0.0230}  \\
			\bottomrule
	\end{tabular}}
 \label{tab:recon}
 \vspace{-4mm}
\end{table}

In Table~\ref{tab:recon}, we compare the PSNR, SSIM, and LPIPS metrics of the original VAE-D8 and our adapted VAE-D4 in reconstructing the images of the Urban100~\cite{huang2015single} dataset. We can see that our VAE-D4 significantly outperforms VAE-D8 in all metrics. Its PSNR is 32.35 dB, showing much better reconstruction quality than VAE-D8, whose PSNR is only 25.29 dB. Furthermore, the SSIM of VAE-D4 is 0.9290, versus 0.7656 for VAE-D8, reflecting improved structural similarity. From the visual example shown in Fig. \ref{fig:introimage1}(a), it is clear that VAE-D4 works much better in preserving image fine-structures than VAE-D8.

\subsection{TVT-based Real-ISR}

\textbf{Training Framework}. With the proposed TVT strategy, our trained VAE-D4 can be directly coupled with the pre-trained denoising UNet in SD for the Real-ISR task. We therefore propose a TVT-based Real-ISR scheme, whose framework is shown in Fig.~\ref{fig:model}(b). Given an LR image $I_{LR}$, we encode it into a latent feature $z_{LR} = E_4(I_{LR})$, and extract the text embedding $c_t$ from it by a prompt extractor, which includes the DAPE~\cite{wu2023seesr} and CLIP text encoder~\cite{rombach2022high}. The restored latent feature $z_{SR}$ is obtained by passing $z_{LR}$ and $c_t$ through the denoising UNet $\epsilon_l$:
\begin{equation}
    z_{SR}=\frac{({z-\sqrt{1-\bar{\alpha}_t} \epsilon_l(z,c_t,t)})}{\sqrt{\bar{\alpha}_t}},
    \label{eq:4}
\end{equation}
where the time step $t$ is set to 1 in our work and $\bar{\alpha}_t$ is the corresponding diffusion coefficient~\cite{ho2020denoising}. Finally, feeding $z_{SR}$ into the decoder $D_4$, we  obtain the SR image $I_{SR}$:
\begin{equation}
    I_{SR}=D_4(z_{SR}).
\end{equation}

To achieve Real-ISR with one-step diffusion, following OSEDiff~\cite{wu2024one}, we add LoRA~\cite{lora} to UNet and use $L_1$ loss, LPIPS loss and VSD loss to train our model:
\begin{equation}
\begin{aligned}
    \mathcal{L}_{SR} & = L_1(I_{SR}, I_{HR}) + \lambda_{1}L_{LPIPS}(I_{SR}, I_{HR}) \\ 
    & + \lambda_{2}L_{VSD}(I_{SR}, I_{HR}),
\end{aligned}
\end{equation}
where $\lambda_{1}$=2, $\lambda_{2}=1$ are weighting hyper-parameters.

\noindent
\textbf{Compute-Efficient UNet}. Our trained VAE-D4 increases the spatial resolution of latent features, which consequently increases the computational demands on the denoising UNet. Previous methods have attempted to alleviate such computation overhead by adjusting channel dimensions~\cite{chen2024adversarial}; however, such a solution requires complete retraining of the UNet, disrupting its pre-trained generative priors. To reduce the computation cost caused by the increase of feature resolution while maintaining pre-trained prior of UNet, we propose a simple yet effective solution: augmenting the UNet with replicated first and last layers (purple blocks, Fig.~\ref{fig:model}(b)) specifically designed to handle high-resolution ($128\times128$) features, while using the original UNet to process the $64\times64$ latent features. We adapt the feature resolution using the default upsampling and downsampling layers from SD 2.1-base. Specifically, we use a $3\times3$ convolutional layer with a stride of 2 to downsample the feature from $128\times128$ to $64\times64$, which is fully trained. Meanwhile, for the upsampling layer, we first employ the nearest neighbor interpolator to upsample the features by a factor of two, followed by a $3\times3$ convolutional layer with a stride of 1 for further mapping of the features, which is also fully trained. In addition, to prevent information loss, skip connection is used to concatenate the $128\times128$ features from the encoder into the decoder.

Our compute-efficient UNet (0.73T FLOPs) design reduces the FLOPs by 51\% compared to directly employing the UNet (1.35T FLOPs) for $128\times128$ resolution inputs. We initialize the replicated layers with the weights of the first and last layers of the original UNet. During training, only the replicated layers undergo full parameter training, while the other UNet layers are fine-tuned using LoRA~\cite{lora}, thereby preserving the integrity of the pre-trained diffusion priors. As we will see in the experimental results, this simple yet effective adaptation allows high-resolution detail recovery without compromising generative capabilities of SD.

\input{table/comparison}

%% file: table/comparison.tex
\begin{table*}[ht]
\caption{Quantitative comparison with SOTA methods on real-world SR task. Symbol `s' denotes the number of diffusion reverse steps. The best and second best results are highlighted in {\color[HTML]{FF0000} \textbf{red}} and {\color[HTML]{6434FC} \textbf{blue}}, respectively.}
\vspace{-3mm}
\centering
	\resizebox{\linewidth}{!}{
\begin{tabular}{l|c|cccccccccccc}
\toprule
Datasets & Metrics       & PSNR↑                                                     & SSIM↑                                  & FSIM↑                                  & LPIPS↓                                 & AFINE-FR↓                                 & DISTS↓                                   & CLIPIQA↑                               & MUSIQ↑                                & Q-Align↑                              & TOPIQ↑                                 & HyperIQA↑ & AFINE-NR↓                              \\ 
\midrule
\multirow{7}{*}{RealSR}      & StableSR-s200 & 24.60                                                     & 0.7047                                 & 0.8296                                 & 0.3068    & -0.7699                    & 0.2163                                                                          & 0.6301                                 & {\color[HTML]{333333} 65.71}          & 3.273                                 & 0.5791                        & 0.5860  & -1.0120                               \\
                             & DiffBIR-s50   & 25.25                                                     & 0.6560                                 & 0.8355                                 & 0.3658             & -0.6771   & 0.2398                                 & {\color[HTML]{6434FC} \textbf{0.6771}} & 66.81                                 & 3.447                                 & 0.6400                                 & 0.6483                          & -0.9265       \\
                             & SeeSR-s50     & 25.38                                                     & 0.7292                                 & 0.8411                                 & 0.2950      & -0.8039                  & 0.2189                                                          & 0.6612                                 & 69.20 & {\color[HTML]{6434FC} \textbf{3.715}} & {\color[HTML]{6434FC} \textbf{0.6800}} & {\color[HTML]{6434FC} \textbf{0.6663}} & -1.0196 \\
                             & PASD-s20      & 25.21                                                     & 0.6798                                 & 0.8430                                 & 0.3380             & {\color[HTML]{6434FC} \textbf{-0.8184}}                    & 0.2260                                                               & 0.6620                                 & {\color[HTML]{6434FC} \textbf{69.77}}                                 & 3.619                                 & 0.6565                                 & 0.6569   & -0.9246                              \\
                             & S3Diff-s1     & 24.38                                                     & 0.7076                                 & 0.8337                                 & {\color[HTML]{6434FC} \textbf{0.2767}}     & -0.8181              & {\color[HTML]{FF0000} \textbf{0.2029}} & {\color[HTML]{333333} 0.6757}          & 67.97                                 & 3.672                                 & 0.6183                                 & 0.6107               & {\color[HTML]{FF0000} \textbf{-1.0589}}                  \\
                             & OSEDiff-s1    & {\color[HTML]{6434FC} \textbf{25.15}}                     & {\color[HTML]{6434FC} \textbf{0.7341}} & {\color[HTML]{6434FC} \textbf{0.8358}} & 0.2921 & -0.7174 & 0.2128                                                               & 0.6693                                 & 69.09                                 & 3.699                                 & 0.6253                                 & 0.6261       & {\color[HTML]{6434FC} \textbf{-1.0489}}                         \\
  \rowcolor{lightpink}     & TVT-s1        & {\color[HTML]{FF0000} \textbf{25.81}}                     & {\color[HTML]{FF0000} \textbf{0.7596}} & {\color[HTML]{FF0000} \textbf{0.8581}} & {\color[HTML]{FF0000} \textbf{0.2587}} & {\color[HTML]{FF0000} \textbf{-0.8787}} & {\color[HTML]{6434FC} \textbf{0.2061}} & {\color[HTML]{FF0000} \textbf{0.6882}} & {\color[HTML]{FF0000} \textbf{69.89}} & {\color[HTML]{FF0000} \textbf{3.770}} & {\color[HTML]{FF0000} \textbf{0.6829}} & {\color[HTML]{FF0000} \textbf{0.6761}} & -1.0237 \\
\midrule
\multirow{7}{*}{DrealSR}     & StableSR-s200 & 28.07                             & 0.7595                                 & 0.8404                                 & 0.3162    & {\color[HTML]{6434FC} \textbf{-0.7320}}                             & 0.2240                                                                  & 0.63095                                & 57.47                                 & 3.062                                 & 0.5116                                 & 0.5258      & -0.8227                           \\
                             & DiffBIR-s50   & 26.55                                                     & 0.6420                                 & 0.8136                                 & 0.4616            & -0.3825                     & 0.2848                                                          & 0.6744                                 & 63.23                                 & 3.314                                 & 0.6514                                 & 0.6410    & -0.7821                             \\
                             & SeeSR-s50     & \color[HTML]{6434FC} \textbf{28.19}                       &  0.7728             & 0.8427             & 0.3133    & -0.5867         & 0.2293                    & 0.6884                                 & 64.68                                 & 3.564                                 & {\color[HTML]{6434FC} \textbf{0.6518}} & {\color[HTML]{6434FC} \textbf{0.6560}} & -0.9143 \\
                             & PASD-s20      & 27.36                                                     & 0.7073                                 & 0.8316                                 & 0.3760     & -0.4164                            & 0.2531                                                              & 0.6808                                 & 61.07                                 & 3.500                                 & 0.6367                                 & 0.6452  & -0.8411                               \\
                             & S3Diff-s1     & 27.09                                                     & 0.7409                                 & 0.8340                                 & 0.3127                       & -0.6226          & {\color[HTML]{FF0000} \textbf{0.2109}} & {\color[HTML]{6434FC} \textbf{0.7127}} & 64.00                                 & {\color[HTML]{6434FC} \textbf{3.607}} & 0.6004                                 & 0.6032            & {\color[HTML]{FF0000} \textbf{-0.9289}}                     \\
                             & OSEDiff-s1    & 27.92                                                     & {\color[HTML]{6434FC} \textbf{0.7835}} & {\color[HTML]{6434FC} \textbf{0.8446}} & {\color[HTML]{6434FC} \textbf{0.2968}} & -0.6656 & {\color[HTML]{6434FC} \textbf{0.2165}}                                                               & 0.6963                                 & {\color[HTML]{6434FC} \textbf{64.65}} & 3.543                                 & 0.6046                                 & 0.6084  & {\color[HTML]{6434FC} \textbf{-0.9178}}                               \\
  \rowcolor{lightpink}     & TVT-s1        & {\color[HTML]{FF0000} \textbf{28.27}}                     & {\color[HTML]{FF0000} \textbf{0.7899}} & {\color[HTML]{FF0000} \textbf{0.8534}} & {\color[HTML]{FF0000} \textbf{0.2900}} & {\color[HTML]{FF0000} \textbf{-0.8057}} & 0.2205 & {\color[HTML]{FF0000} \textbf{0.7220}} & {\color[HTML]{FF0000} \textbf{65.56}} & {\color[HTML]{FF0000} \textbf{3.641}} & {\color[HTML]{FF0000} \textbf{0.6591}} & {\color[HTML]{FF0000} \textbf{0.6655}} & -0.9073 \\
\midrule
\multirow{7}{*}{Div2k-val}  & StableSR-s200 & 23.28                                                     & 0.5728                                 & 0.8182                                 & 0.3108  & -0.7484                     & 0.2044                                   & 0.6766                                 & 65.93                                 & 3.147                                 & 0.5991                                 & 0.6138   & -0.8872                              \\
                             & DiffBIR-s50   & 23.39                                                     & 0.5441                                 & 0.8134                                 & 0.3762  & -0.4793                   & 0.2284                                                            & 0.6745                                 & 66.37                                 & 3.518                                 & 0.6611                                 & 0.6427   & -0.8765                      \\
                             & SeeSR-s50     & {\color[HTML]{6434FC} \textbf{23.82}}                                                     & {\color[HTML]{6434FC} \textbf{0.6115}}                                 & 0.8232                                 & 0.3149    & -0.8105                    & 0.1941                                                                  & 0.6800                                 & 68.01 & {\color[HTML]{FF0000} \textbf{4.129}} & {\color[HTML]{6434FC} \textbf{0.6745}} & {\color[HTML]{333333} 0.6604}     & -0.9450     \\
                             & PASD-s20      & 23.14                                                     & 0.5505                                 & 0.8082                                 & 0.3571     & -0.5330                & 0.2207                                                              & 0.6788                                 & {\color[HTML]{FF0000} \textbf{68.95}} & 3.795                                 & 0.6594                                 & {\color[HTML]{6434FC} \textbf{0.6630}} & -0.9239 \\
                             & S3Diff-s1     & 23.00                                                     & 0.5800                                 & 0.8233                                 & {\color[HTML]{FF0000} \textbf{0.2606}} & {\color[HTML]{FF0000} \textbf{-1.1101}} &{\color[HTML]{FF0000} \textbf{0.1739}} & {\color[HTML]{FF0000} \textbf{0.7007}} & 67.88                                 & {\color[HTML]{333333} 3.867}          & 0.6341                                 & 0.6245   & {\color[HTML]{FF0000} \textbf{-0.9868}}                \\
                             & OSEDiff-s1    & 23.72                     & 0.6108 & {\color[HTML]{6434FC} \textbf{0.8246}} & 0.2941 & -0.8388                                & 0.1976                          & 0.6683                                 & 67.97                                 & 3.771                                 & 0.6188                                 & 0.6374 & {\color[HTML]{6434FC} \textbf{-0.9200}}                                \\
  \rowcolor{lightpink}     & TVT-s1        & {\color[HTML]{FF0000} \textbf{24.23}}                     & {\color[HTML]{FF0000} \textbf{0.6292}} & {\color[HTML]{FF0000} \textbf{0.8410}} & {\color[HTML]{6434FC} \textbf{0.2773}} & {\color[HTML]{6434FC} \textbf{-0.9132}} & {\color[HTML]{6434FC} \textbf{0.1860}}                                  & {\color[HTML]{6434FC} \textbf{0.6986}} & {\color[HTML]{6434FC} \textbf{68.67}}                                 & {\color[HTML]{6434FC} \textbf{3.920}} & {\color[HTML]{FF0000} \textbf{0.6791}} & {\color[HTML]{FF0000} \textbf{0.6794}} & -0.8966 \\ 
 \bottomrule
 \end{tabular}}
    \label{tab:results}
	\vspace{-4mm}
\end{table*}

%% file: sec/4_exp.tex
\section{Experiments}

\subsection{Experimental Settings}

\noindent
\textbf{Training Datasets.} To train the VAE, we utilize the many images in OpenImage dataset~\cite{kuznetsova2020open} and an additional 1.44 million 512×512 high-quality image patches from the LSDIR dataset~\cite{li2023lsdir} as training data. For the Real-ISR task, following previous SD-based methods~\cite{wang2023exploiting, wu2023seesr,yang2023pixel,wu2024one}, we employ the LSDIR dataset~\cite{li2023lsdir} and the first 10K face images from the FFHQ dataset~\cite{ffhq} as the training data. The Real-ESRGAN degradation pipeline~\cite{wang2021real} is used to generate LR-HR training pairs. 

In addition, to more comprehensively validate the fine-structure preservation capability of our TVT method, we further conduct experiments on the RealCE dataset~\cite{ma2023benchmark}, a benchmark specifically designed for real-world scene text image super-resolution (STISR). The RealCE dataset provides 1,935 LR-HR image pairs for training, along with 261 testing pairs for $4\times$ zoom. It features a total of 24,666 Chinese text lines and 9,123 English text lines, accommodating various sizes. In this experiment, we utilize the RealCE training set as our training data.

\noindent
\textbf{Testing Datasets.}  
To demonstrate the performance of our proposed method on Real-ISR, we employ the test datasets used in previous works~\cite{wang2023exploiting, wu2024one}, including DIV2K-val~\cite{div2k}, RealSR~\cite{realsr} and DRealSR~\cite{drealsr}. The DIV2K-val dataset consists of 3,000 images of $512\times512$ resolution, with LR-HR pairs generated using the Real-ESRGAN degradation pipeline, while RealSR and DRealSR are datasets of real-world LR-HR pairs. Furthermore, we use the RealCE dataset~\cite{ma2023benchmark} to test our method in STISR.

\noindent
\textbf{Compared Methods.} For the Real-ISR task, we compare our proposed TVT method with current advanced SD-based Real-ISR methods, including multi-step diffusion methods (StableSR~\cite{wang2023exploiting}, PASD~\cite{yang2023pixel}, DiffBIR~\cite{lin2023diffbir} and SeeSR~\cite{wu2023seesr}) and one-step diffusion methods (OSEDiff~\cite{wu2024one} and S3Diff~\cite{zhang2024degradation}). For the real-world STISR task, we compare TVT with a variety of methods, including text-prior networks~\cite{wang2020scene, ma2022text, ma2023text, chen2021scene}, diffusion-based methods~\cite{singh2024dcdm}, and SD-based methods~\cite{zhang2024diffusion, yang2023pixel, wu2024one}. Note that PASD~\cite{yang2023pixel} and OSEDiff~\cite{wu2024one} are re-trained on the RealCE dataset.

\noindent
\textbf{Evaluation Metrics.} For the Real-ISR task, we utilize a set of reference-based and no-reference metrics to evaluate the competing methods. The reference-based metrics include PSNR, SSIM~\cite{ssim}, FSIM~\cite{zhang2011fsim}, LPIPS~\cite{lpips}, DISTS~\cite{dists}, and the fidelity part of the A-FINE~\cite{chen2025toward} model, denoted as AFINE-FR. Note that PSNR and SSIM are computed on the Y channel of YCbCr space. The no-reference metrics include CLIPIQA~\cite{clipiqa}, MUSIQ~\cite{musiq}, Q-Align~\cite{wu2023q}, TOPIQ~\cite{chen2024topiq}, HyperIQA~\cite{su2020blindly}, and the naturalness part of the A-FINE~\cite{chen2025toward} model, denoted as AFINE-NR.
For the STISR task, the employed reference-based metrics include PSNR, SSIM~\cite{ssim} and LPIPS~\cite{lpips}. In addition, the character  recognition accuracy~(ACC) and normalized edit distance~(NED) are used to demonstrate the quality of text reconstruction.

\begin{figure*}[th]
	\centering
	\includegraphics[width=0.8\linewidth]
	{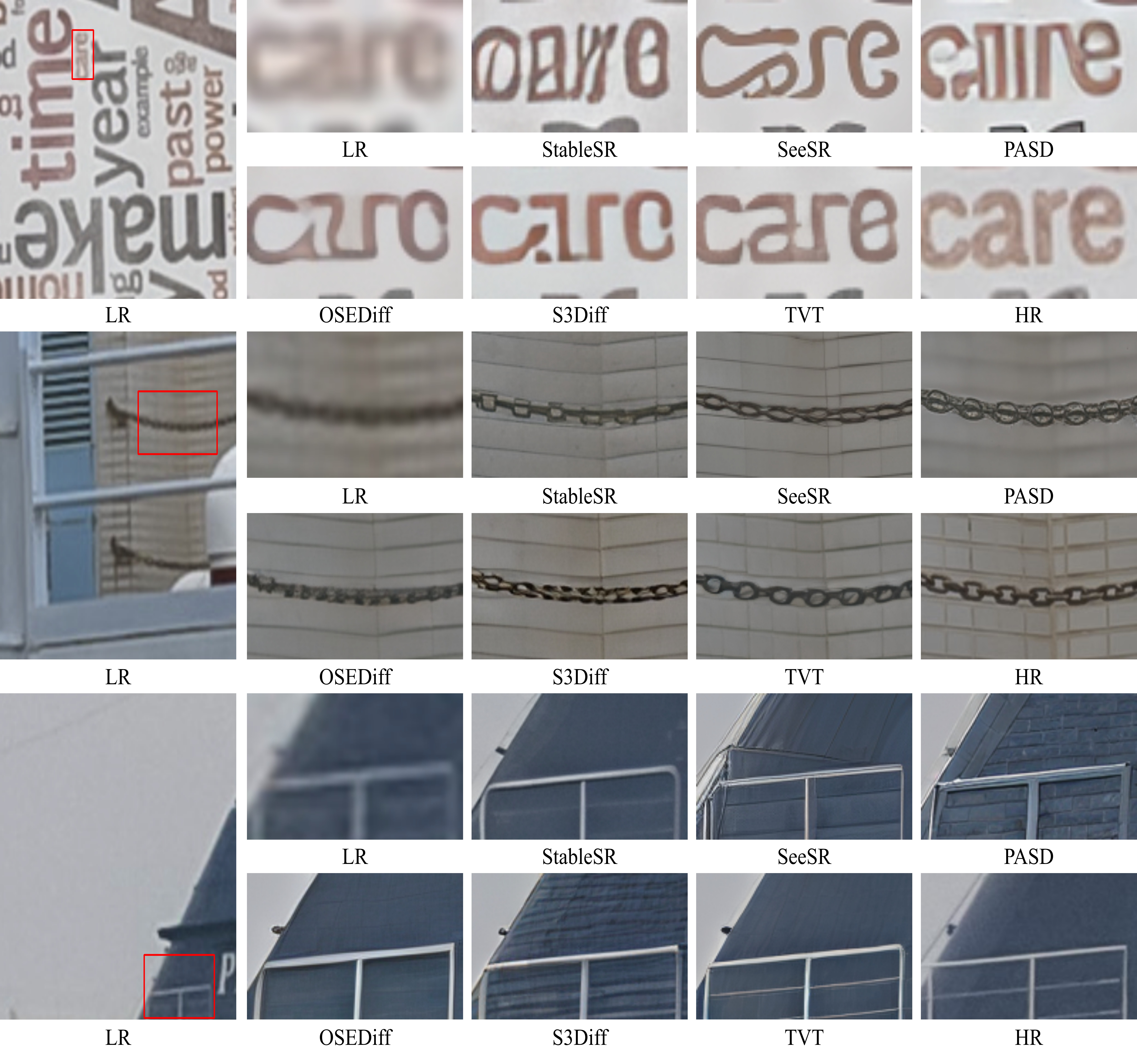}
        \vspace{-4mm}
	\caption{Visual comparison with SD-based Real-ISR methods. Please zoom in for a better view.}
	\label{fig:result1}
    \vspace{-5.5mm}
\end{figure*}

\noindent
\textbf{Implementation Details.} The VAE-D4 encoder and decoder are trained using a batch size of 256 for $2\times10^5$ iterations on 8 A100 GPUs. For both Real-ISR and STISR experiments, we utilize the SD 2.1-base \cite{rombach2022high} as the pre-trained diffusion model. Our proposed TVT-based Real-ISR method is trained on $512 \times 512$ images, employing a learning rate of $5 \times 10^{-5}$ and a batch size of 16. The AdamW optimizer~\cite{adamw} is used for optimization. The rank of the LoRA utilized in our TVT model training is set to 4. The training process encompasses $2\times10^4$ iterations on 4 A100 GPUs in the Real-ISR experiment, and $1 \times 10^4$ iterations in the STISR experiment. 

\subsection{Comparison with State-of-the-Arts}

\begin{table}
	\caption{Quantitative comparison with SOTA methods on real-world STISR task. The best and second best results are highlighted in {\color[HTML]{FF0000} \textbf{red}} and {\color[HTML]{6434FC} \textbf{blue}}, respectively. Symbol `*' means that the method is re-trained with edge-aware loss\cite{ma2023benchmark}.}
            \vspace{-3mm}
	\centering
	\resizebox{\linewidth}{!}{
		\begin{tabular}{c|ccccc}
			\toprule
			Methods & PSNR↑ & SSIM↑ & LPIPS↓ & ACC↑ & NED↑ \\ \midrule
			TSRN~\cite{wang2020scene} & 18.11 & 0.4850 & 0.1981 & 0.2316 & 0.4159 \\
       TPGSR~\cite{ma2023text} & 18.07 & 0.4758 & 0.1843 & 0.2326 & 0.4123 \\
       TBSRN~\cite{chen2021scene} & 18.33 & 0.4826 & 0.1715 & 0.2527 & 0.4444\\
       TATT~\cite{ma2022text} & 17.96 & 0.4904 & 0.1804 & 0.2330 & 0.4342 \\ 
       TATT*~\cite{ma2023benchmark} & 17.89 & 0.4822 & 0.1546 & 0.2417 & 0.4549  \\ 
       \midrule
       DCDM~\cite{singh2024dcdm} & 18.91 & 0.5090 & N/A & 0.2549 & N/A\\
       DiffTSR~\cite{zhang2024diffusion} & 18.85 & 0.6533 & 0.1725 & 0.2494 & 0.4657\\
       PASD~\cite{yang2023pixel} & 19.34 & 0.6724 & 0.2003 & 0.2914 & {\color[HTML]{6434FC} \textbf{0.6413}} \\
       OSEDiff~\cite{wu2024one} & {\color[HTML]{FF0000} \textbf{19.89}} & {\color[HTML]{6434FC} \textbf{0.7003}} & {\color[HTML]{6434FC} \textbf{0.1707}} & {\color[HTML]{6434FC} \textbf{0.2998}} & 0.6400 \\
 \rowcolor{lightpink}      TVT &{\color[HTML]{6434FC} \textbf{19.69}}&{\color[HTML]{FF0000} \textbf{0.7030}}&{\color[HTML]{FF0000} \textbf{0.1644}}&{\color[HTML]{FF0000} \textbf{0.3096}}&{\color[HTML]{FF0000} \textbf{0.6621}}\\
	\bottomrule
	\end{tabular}}
 \label{tab:stisr}
 \vspace{-4mm}
\end{table}

\textbf{Quantitative Comparison.} Table \ref{tab:results}  quantitatively compares our proposed TVT method with other SD-based methods on the Real-ISR task. One can see that TVT demonstrates superior performance across all the three datasets. We can have the following observations. (1) For reference-based metrics, TVT consistently achieves the best PSNR/SSIM/FSIM scores, and the best or second best LPIPS/AFINE-FR/DISTS scores. 
Such an improvement comes from the higher fidelity of our VAE-D4, enabling TVT to have better ability to protect the image fine-structures. For example, on RealSR, the PSNR and SSIM indices of TVT (25.81/0.7596) are significantly higher than its competitors S3Diff (24.38/0.7076) and OSEDiff(25.15/0.7341). (2) For non-reference metrics, TVT achieves the most counts of best or second best scores on all the datasets. Note that this is non-trivial, as most SD-based methods need to make a trade-off between reference and no-reference metrics. The superior results of TVT on both reference and no-reference metrics validate that, through fine-structure preservation, TVT can generate more faithful and visually pleasing image details.

The quantitative comparison on the RealCE dataset is presented in Table \ref{tab:stisr}. One can see that TVT achieves the best (SSIM/LPIPS) or second best (PSNR) results on reference-based image quality metrics, surpassing its SD and even non-SD based competitors. Meanwhile, TVT is significantly better than other methods in recognition accuracy and NED indicators. These results validate not only TVT's superior ability in restoring high-quality images but also its advantages in preserving structural information, which is important for character recognition.

\begin{table}
\caption{Comparison of parameter count (Para.) and FLOPs.}
 \vspace{-3mm}
	\centering
	\resizebox{\linewidth}{!}{
		\begin{tabular}{c|cccccc>{\columncolor{lightpink}}c}
			\toprule
			Methods & StableSR & DiffBIR & SeeSR & PASD & S3Diff & OSEDiff & TVT \\ \midrule
        Steps &200&50&50&20&1&1&1\\
	Para. (B) &1.56&1.68&2.51&2.31&1.33&1.77&1.72\\
        FLOPs (T) &79.94&24.31&65.86&29.13&2.62&2.27&1.97\\
	\bottomrule
	\end{tabular}}
 \label{tab:flops}
 \vspace{-6mm}
\end{table}

\begin{figure}
	\centering
	\includegraphics[width=1\linewidth]
	{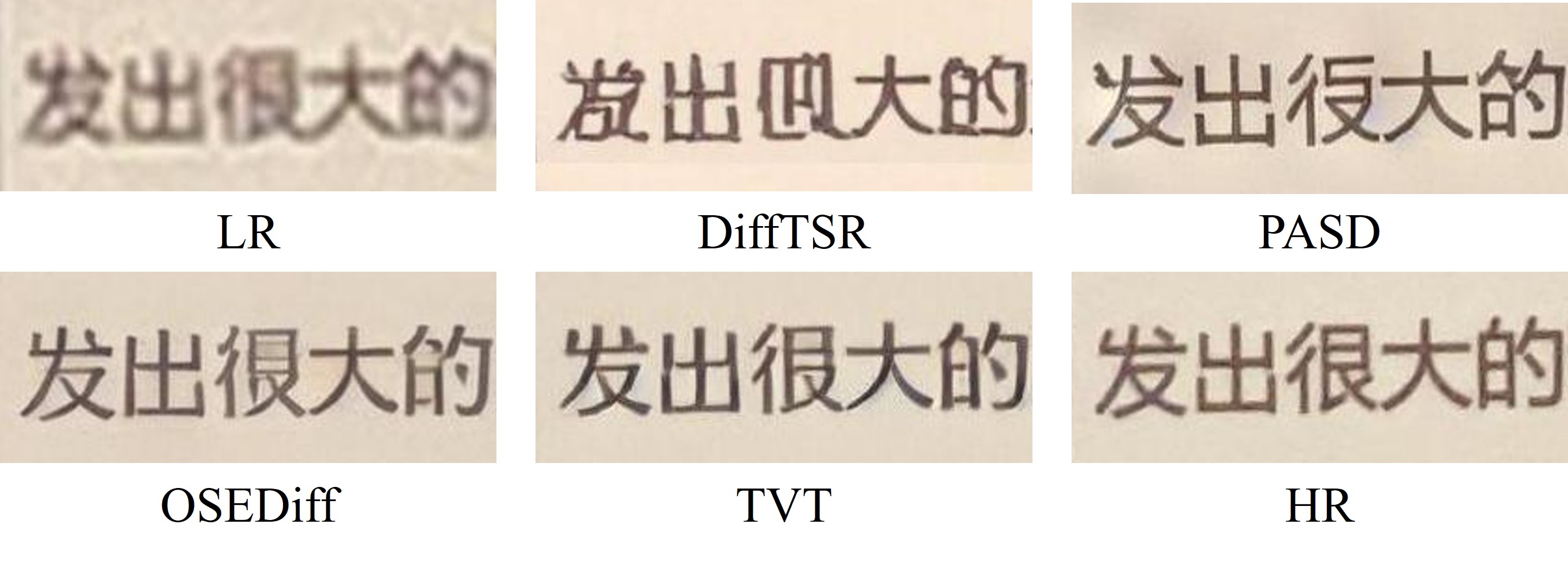}
        \vspace{-8mm}
	\caption{Visual comparison with STSR methods.}
	\label{fig:tvtresult}
    \vspace{-6mm}
\end{figure}

\noindent
\textbf{Qualitative Comparison.} The visaul comparisons are shown in Fig.~\ref{fig:result1}. It is evident that TVT significantly outperforms other methods on structure preservation. Although the competing methods can improve image quality over the LR input, they struggle with preserving the fine structures due to the limitation of VAE-D8. For example, both S3Diff and OSEDiff fail to reconstruct the text `care' in the first image. In contrast, our proposed TVT method presents sharper and more legible text output, illustrating its effectiveness in reproducing image details with high fidelity. In addition, SeeSR and PASD exploit high-level information to generate semantically consistent images, which may cause some erroneous structures (\eg, incorrect text/chain in the first/second images) or over-enhanced structures (\eg, the unnatural structure in the third image). Fig. \ref{fig:tvtresult} illustrates the visual results on the STISR task. It can be seen that TVT can restore the text structure much better than the other methods. More visualization comparisons can be found in the \textbf{supplementary materials}.

\noindent
\textbf{Complexity Comparisons}. Table \ref{tab:flops} compares the SD-based Real-ISR methods in terms of parameter count (Para.) and floating-point operations per second (FLOPs). Despite the increased latent feature resolution introduced by our VAE-D4, our compute-efficient UNet reduces much the computational complexity. Specifically, TVT achieves the lowest FLOPs at 1.97T, and it has 1.72B parameters, fewer than OSEDiff. Overall, our TVT strategy, together with efficient design of VAE-D4 and denoising UNet, makes our Real-ISR model effective and efficient.

\subsection{Ablation Study}

Since it is time-consuming to train a VAE, in the ablation experiments, we train the VAEs only on the OpenImage dataset to accelerate the experimental process. Therefore, the TVT results in the ablation study may differ slightly from those reported in Table \ref{tab:results}.

\noindent
\textbf{The Setting of Compact VAE-D4.}  
Table \ref{tab:ablationvae} illustrates the impact of various VAE-4D configurations on Real-ISR performance. Variant `V1' utilizes three hierarchical stages (128, 256, 512 channels) along with a middle block. In contrast, V2 employs a simplified VAE-D4 consisting solely of the three hierarchical stages (128, 256, 512 channels). As shown in Table \ref{tab:ablationvae}, TVT achieves similar performance to both V1 and V2, while requiring fewer FLOPs.

\noindent
\textbf{TVT Strategy.} To validate the effectiveness of the our TVT strategy, we conduct three ablation experiments, as shown in Table \ref{tab:ablationtvt}. In T1, we train a VAE-D4 from scratch without employing the TVT strategy. In contrast, T2 trains a VAE-D4 while using $L_1$ loss to align the latent feature of VAE-D4 with that of VAE-D8. As shown in Table \ref{tab:ablationtvt}, compared to TVT and T2, the performance of T1 significantly degrades in terms of PSNR, SSIM, CLIPIQA and MUSIQ. This indicates that if the latent space is not aligned with that of VAE-D8, the VAE-D4 cannot leverage pre-trained UNet's generative capabilities. Furthermore, compared to the TVT strategy, aligning VAE-D4 and VAE-D8 by $L_1$ loss will also significantly degrades the performance. This highlights the effectiveness of the TVT strategy in aligning the the latent spaces of VAE-D4 and VAE-D8.

\begin{table}
	\centering
        \caption{Ablation studies on the setting of compact VAE-D4 on RealSR dataset.}
	\resizebox{0.85\linewidth}{!}{
		\begin{tabular}{c|cccccc}
			\toprule
			Methods & FLOPs & PSNR↑ & SSIM↑ & CLIPIQA↑ & MUSIQ↑ \\ \midrule
       V1 & 3.02 T &25.84&0.7620&0.6934&69.56\\
       V2 & 2.40 T &25.77&0.7603&0.6930&69.31\\
     \rowcolor{lightpink}  TVT & 1.97 T &25.83&0.7619&0.6924&69.26\\
	\bottomrule
	\end{tabular}}
 \label{tab:ablationvae}
 \vspace{-2mm}
\end{table}

\begin{table}
	\centering
        \caption{Ablation studies on TVT strategy on  RealSR dataset.}
	\resizebox{0.85\linewidth}{!}{
		\begin{tabular}{c|cccc}
			\toprule
			Methods & PSNR↑ & SSIM↑ & CLIPIQA↑ & MUSIQ↑ \\ \midrule
       T1 (w/o TVT) & 25.50 & 0.7549 & 0.6320 & 67.85 \\
       T2 (w/ $L_1$ loss) & 25.75 & 0.7602 & 0.6734 & 69.10 \\
  \rowcolor{lightpink}     TVT & 25.83 & 0.7619 & 0.6924 & 69.26 \\
	\bottomrule
	\end{tabular}}
 \label{tab:ablationtvt}
 \vspace{-2mm}
\end{table}

\begin{table}
	\centering
        \caption{Ablation studies on TVT-based Real-ISR method on RealSR dataset. VAE-D8-SC is the VAE-D8 with skip-connection. VAE-D8C8 means the $8\times$ downsampled VAE with eight channels. Prun-UNet is a UNet that has pruned 25\% of the channels from SD's UNet. CE-UNet is our compute-efficient UNet.}
        \resizebox{0.9\linewidth}{!}{
		\begin{tabular}{l|cccc}
			\toprule
			Methods & FLOPs & SSIM↑ & MUSIQ↑ & Q-Align↑ \\ \midrule
       S1 (VAE-D8+UNet) & 2.27 T & 0.7419 & 68.86 &3.592\\
      S2 (VAE-D8-SC+UNet) & 2.27T & 0.7504 & 68.12 & 3.482\\
       S3 (VAE-D8C8+UNet) & 2.27 T & 0.7456 & 67.79 & 3.514\\
       S4 (VAE-D4+UNet) & 2.59 T & 0.7632 & 69.40 & 3.797 \\
       S5 (VAE-D4+Prun-UNet) & 2.00 T & 0.7446 & 68.03 & 3.4357 \\
\rowcolor{lightpink}       TVT (VAE-D4+CE-UNet) & 1.97 T & 0.7619 & 69.26 & 3.770\\
	\bottomrule
	\end{tabular}}
 \label{tab:ablationunet}
 \vspace{-2mm}
\end{table}

\noindent
\textbf{The Setting of TVT-based Real-ISR Method.} 
We conduct six ablation experiments to demonstrate the effectiveness of VAE-D4 and compute-efficient UNet (CE-UNet) for Real-ISR, as shown in Table \ref{tab:ablationunet}. S1 uses VAE-D8 and pre-trained UNet in SD 2.1-base. S2 employs an VAE-D8 with skip connections~(VAE-D8-SC) to protect the structure information. In S3, we train a $8\times$ downsampled VAE with eight channels of latent features~(VAE-D8C8), whose latent space is aligned with VAE-D8 by interpolating the number of channels in VAE-D8. S4 employs our proposed compact VAE-D4 and pre-trained SD UNet. In S5, we reduce the FLOPs of pre-trained UNet by pruning 25\% of it's channels (Prun-UNet). Comparing S2 with S1, we can find that skip connections cannot effectively improve the fidelity and perceptual quality. The main issue is that the VAE encoder features are much degraded, which hinder VAE decoder’s reconstruction, making the model struggle to balance fidelity and perceptual quality. Meanwhile, compared with S1, the SSIM of S3 is improved to 0.7456 while the MUSIQ and Q-Align of S2 drop to 67.79 and 3.51, respectively. This indicates that increasing the number of feature channels can alleviate information loss in VAE-D8. However,  increasing the number of feature channels may disrupt the original prior of the U-Net, thus affecting the generative ability of SD. In addition, comparing S4 with S1, we see that VAE-D4 significantly improves in SSIM, MUSIQ and Q-Align by 0.0202, 0.49 and 0.198, respectively. Furthermore, compared to S4 and S5, our proposed CE-UNet significantly reduces FLOPs while achieving comparable performance to S4 in SSIM, MUSIQ, and Q-Align. This shows that CE-UNet effectively preserves the pre-training priors of SD UNet while reducing FLOPs.

%% file: sec/5_con.tex
\section{Conclusion}

We presented a Transfer VAE Training (TVT) approach for fine-structure preserved Real-ISR by adapting the $8\times$ downsampled VAE (VAE-D8) of SD into a $4\times$ downsampled variant (VAE-D4) while keeping the latent space nearly unchanged. Through a two-stage training framework, which first aligned the VAE-D4 decoder with the VAE-D8 latent space and then adapted VAE-D4 encoder with the aligned decoder, we seamlessly transferred the VAE-D4 into the pre-trained SD latent space without costly UNet retraining. Meanwhile, the VAE-D4 was compactly designed, and a compute-efficient UNet was proposed to efficiently process the higher resolution features introduced by VAE-D4. Extensive experiments demonstrated that our TVT-based Real-ISR method outperformed state-of-the-art SD-based Real-ISR approaches in preserving image fine-structures with even lower FLOPs.

\begin{figure}[t]
    \centering
    \vspace{-3mm}
    \includegraphics[width=1\linewidth]{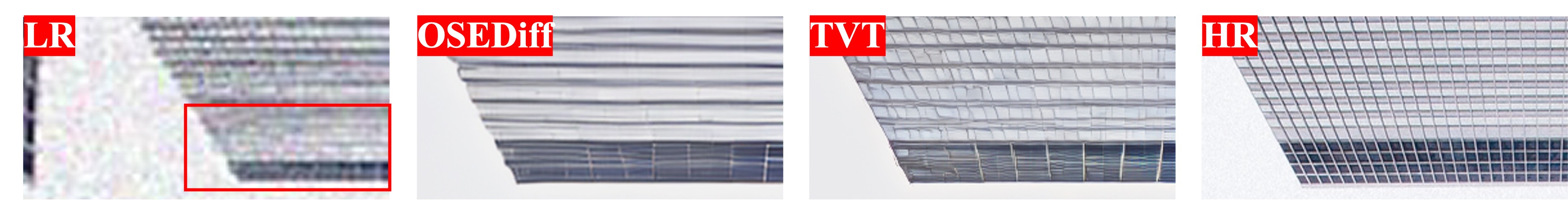}
    \vspace{-6mm}
    \caption{Our method faces challenges in recovering details when fine-scale structures are severely degraded.}
    \label{fig:challenge}
    \vspace{-5mm}
\end{figure}

\noindent 
\textbf{Limitations}. While we have designed a compact VAE and a compute-efficient UNet to mitigate the computational cost brought by the higher resolution features of our VAE-D4, it still faces challenges on resource-constrained devices, which demand models with fewer parameters and lower FLOPs. 
In addition, although our trained VAE-D4 significantly improves the fine-structure details, it still faces challenges when fine-scale structures are severely degraded. One example is shown in Fig.~\ref{fig:challenge}.